# Hybrid Motion Planning Task Allocation Model for AUV's Safe Maneuvering in a Realistic Ocean Environment

Somaiyeh MahmoudZadeh[1], David M.W Powers[1], Karl Sammut[2], Amir Mehdi Yazdani[2,], Adham Atyabi[3]

[1] College of Science and Engineering, Flinders University, Adelaide, SA, Australia
[2] Centre for Maritime Engineering, Control and Imaging, Flinders University, Adelaide, SA, Australia
[3] Seattle Children's Research Institute, University of Washington, United States

somaiyeh.mahmoudzadeh@flinders.edu.au
david.powers@flinders.edu.au
karl.sammut@flinders.edu.au
amirmehdi.yazdani@flinders.edu.au
adham.atyabi@seattlechildrens.org

**Abstract** This paper presents a hybrid route-path planning model for an Autonomous Underwater Vehicle's task assignment and management while the AUV is operating through the variable littoral waters. Several prioritized tasks distributed in a large scale terrain is defined first; then, considering the limitations over the mission time, vehicle's battery, uncertainty and variability of the underlying operating field, appropriate mission timing and energy management is undertaken. The proposed objective is fulfilled by incorporating a route-planner that is in charge of prioritizing the list of available tasks according to available battery and a path-planer  that acts in a smaller scale to provide vehicle's safe deployment against environmental sudden changes. The synchronous process of the task assign-route and path planning is simulated using a specific composition of Differential Evolution and Firefly Optimization (DEFO) Algorithms. The simulation results indicate that the proposed hybrid model offers efficient performance in terms of completion of maximum number of assigned tasks while perfectly expending the minimum energy, provided by using the favorable current flow, and controlling the associated mission time. The Monte-Carlo test is also performed for further analysis. The corresponding results show the significant robustness of the model against uncertainties of the operating field and variations of mission conditions.

**Nomenclature**

| | | | |
|---|---|---|---|
| $\aleph_i$ | Task index | $\Gamma_{3\text{-}D}$ | Symbol of the three dimensional terrain |
| $\rho_i$ | Priority of task $i$ | $\eta$ | The AUV state on NED frame $\{n\}$ |
| $\xi_i$ | Risk percentage associated with task $i$ | $[X,Y,Z]$ | Vehicles North, $x$, East, $y$, Depth, $z$, position along the path $\wp$ |
| $\delta_i$ | Absolute time required for completion of task $i$ | $\phi$ | The Euler angle of roll |
| $P$ | Vertices of the network that corresponds to waypoints | $\theta$ | The Euler angle of pitch |
| $E$ | Edges of the network | $\psi$ | The Euler angle of yaw |
| $m$ | Number of waypoints in the network | $\upsilon$ | Vehicle's water referenced velocity in the body frame $\{b\}$ |
| $k$ | Number of edges in the network | $u$ | The surge component of the velocity $\upsilon$ |
| $p^i_{x,y,z}$ | Position of arbitrary waypoint $i$ in 3-D space | $v$ | The sway component of the velocity $\upsilon$ |
| $e_{ij}$ | An arbitrary edge that connects $p^i_{x,y,z}$ to $p^j_{x,y,z}$ | $w$ | The heave component of the velocity $\upsilon$ |
| $w_{ij}$ | The weight assigned to $e_{ij}$ | $\wp$ | The potential trajectory generated by the local path planner |
| $d_{ij}$ | Distance between position of $p^i_{x,y,z}$ and $p^j_{x,y,z}$ | $\vartheta$ | Control point along the path $\wp$ |
| $t_{ij}$ | Time required for traversing edge $e_{ij}$ | $n$ | Number of control points along an arbitrary path $\wp$ |
| $\Theta$ | Obstacle | $L_\wp$ | Length of the candidate path $\wp$ |
| $\Theta_p$ | Obstacle's position | $T_\wp$ | The local path flight time |
| $\Theta_r$ | Obstacle's radius | $T_{exp}$ | The expected time for passing an edge |
| $\Theta_{Ur}$ | Obstacle's uncertainty rate | $\wp_{CPU}$ | computational time for generating a local path |
| $V_C$ | The current velocity vector | $\Re$ | An arbitrary route including sequences of tasks and waypoints |
| $u_c$ | X component of the current vector | $T_\Re$ | The route travelled time |
| $v_c$ | Y component of the current vector | $T_\tau$ | The total available time for the mission |
| $S$ | Two dimensional x-y space | $T_{compute}$ | Computation time for checking re-routing criterion and its process |
| $S^o$ | The center of the vortex in the current map | $C_\wp$ | The cost of local path generated by path planner |
| $\ell$ | The radius of the vortex in the current map | $C_\aleph$ | The cost of tasks completion |
| $\Im$ | The strength of the vortex in the current map | $C_\Re$ | The total cost of route including $C_\wp$ and $C_\aleph$ |

## 1 Introduction

Autonomous Underwater Vehicles (AUVs) have been discovered as the most cost-effective and expedient technology in carrying out the underwater missions over the past and coming years. They are largely employed for various purposes such as scientific underwater explorations [1], inspection and surveys [2], sampling and monitoring coastal areas [3], offshore installations and mining industries [4], etc. However, most of the available AUVs operate with a pre-programmed mission scenario while all parameters for entire mission should be defined in advance and operator's interaction is necessary issue. For any of scientific, surveillance, mine or military applications of the AUV, a sequence of tasks is predefined and fed to vehicle in the series of commands format (mission scenario) that limits the mission to executing a list of pre-programmed instructions and completing a predefined sequences of tasks. Hence, an advanced level of autonomy is an essential prerequisite to trade-off within importance of tasks and problem restrictions while adapting the terrain changes during the operation, in which having a robust motion planning strategy and accurate task allocation scheme are substantial requirements in this regard. Motion planning and vehicle task allocation on different frameworks have been comprehensively investigated over the past two decades. Various deterministic and heuristic strategies have been suggested for unmanned vehicles' path/trajectory planning such as D* [5], A* [6,7], Fast Marching (FM) algorithm [8], and FM*[9]. Cui et al., (2016) proposed an adaptive Mutual information-based path planning algorithm for multi-AUV operations [10]. This approach used multidimensional RRT* to estimate the scalar field sampling over a region of interest where the estimated sampling positions get improved by maximizing the mutual information between the observations and scalar field model [10]. The well-known direct method of optimal control theory, called inverse dynamics in the virtual domain (IDVD) method, was employed to develop and test a real-time trajectory generator for realization on board of an AUV [11]. Vehicle routing and task scheduling problem also has been vastly studied in recent years and many strategies have been suggested such as graph matching algorithm [12], Tabu search algorithm [13], partitioning method [14], simulated annealing [15], and branch and cut algorithm [16]. Assuming that the tasks for a specific mission are distributed in different areas of a waypoint cluttered graph-like terrain, there should be a compromise among prioritizing the tasks according to available battery/time in a way that vehicle is guided toward the destination waypoint, which is combination of a discrete and a continuous optimization problem at the same time. Hence, the vehicle task allocation-routing is categorized as a Non-deterministic Polynomial-time (NP) hard problem due to its combinatorial nature, which is analogous to both Knapsack and Traveler Salesman Problems (TSP). The time efficient path planning is also an NP-Hard problem often solved by optimization algorithms. The computational burden is overshadowed by increment of the problem search space (e.g complexity of the graph topology or terrain vastness), which is an intricate issue and should be taken into consideration. The deterministic and heuristic methods are computationally time consuming that has a detrimental effect on real-time performance of the motion planning problem; hence, these algorithms are not suitable for real-time applications. Bio-inspired meta-heuristic optimization algorithms are diverse nature inspired algorithms and are known as new revolution in solving complex and hard problems. These algorithms are the fastest approach presented for solving NP-hard complexity of motion planning problems and are capable of producing near optimal solutions [17], which is appropriate for the purpose of this study.

*Meta-heuristic Optimization Algorithm: The State of the Art in Motion Planning and Task Assignment*

Meta-heuristics are cost based non-deterministic optimization algorithms that mimic the nature to efficiently solve the complex problems. Former methods to solve motion planning problems (discussed above) require considerable computational efforts and tend to fail when the problem size grows. A vast literature exists on evolutionary or swarm based optimization algorithms for solving both vehicle's optimum path planning, routing and task scheduling problems. Despite meta-heuristics do not necessarily produce pure optimal solutions, but they are computationally fast and efficient and especially appropriate for the real-time applications [18, 19]. The Particle Swarm Optimization (PSO) [18, 20] and Quantum-based PSO (QPSO) [21] are two swarm-based optimization methods applied successfully on AUV path planning problem. An offline three-dimensional path planner based on a non-dominated sorting genetic algorithm (NSGA-II) is proposed for waypoint guidance of an AUV [22]. A Differential Evolution (DE) based path planner is applied to an AUV path planning in a severe underwater environment [23]. In the scope of routing-task-assigning also, a time-optimal conflict free route planning relying on an adaptive Genetic Algorithm (GA) is proposed by Kwok et al., [12] that could facilitate the AUV to operate in a large-scale sea terrain with a few waypoints. An evolution based AUV route planner has been developed by MahmoudZadeh et al., [24] in which the vehicle's operation is considered in a large scale static network of waypoints and two GA and PSO algorithms have been applied to solve the graph complexity of the routing problem. Subsequently, their proposed method was extended to more complex environment encountering a semi-dynamic operation network and efficiency of two other evolutionary algorithms of Biogeography-based Optimization (BBO) and PSO were tested and compared on vehicle's dynamic task assignment and routing [25]. Certainly, having a more efficient optimization approach for solving vehicle routing and path planning problems to achieve faster CPU time and competitive performance is still an open area for research. Additional to importance of the employed algorithm, another difficulty of large scale operations is challenges associated with the behavior of a dynamic uncertain terrain that cause a pre-planned trajectory becomes inefficient or even invalid over time. On the other hand, the path planning strategies are not provided for handling vehicle's task assignment, specifically in cases that several tasks are distributed in a waypoint cluttered graph-like terrain where vehicle is required to carry out a specific sequence of prioritized tasks. Thus, a routing strategy is required to handle graph search constraints and carrying out the task assignment. With respect to above discussion, existing approaches mainly are able to cover only a part of this problem either task assignment together with time management or path planning with safety considerations. To address both algorithmic and technical problems associated with large and small scale motion planning, this study constructs a combinatorial DEFO framework including a higher level task-allocate-routing strategy along with a small scale evolution based path planner.

*Research Contribution*

The subject area is one that is of importance, high-level mission planning of AUVs in the field is still far from automated, drawing on judgement of experienced human operators. Steps to reduce the reliance on expert operators would contribute to scalability of AUV operations and also improve reliability and repeatability of operations. The main contribution of this paper is joining two disparate prospective of vehicle's autonomy in high level task organizing and low level motion planning in a real-time manner. The system is advantageous due to having a cooperative and concordant manner in which adopting diverse algorithms by these modules do not detriment the real-time performance of the system. The total motion of AUV for a mission is described a thread of trajectory from the start waypoint to the goal waypoint by passing through all the necessary prioritized waypoints. The High level is capable of finding a time efficient route and appropriate arrangement of tasks to ensure the AUV has a plenteous journey and efficient timing. The path planner, in this context, is responsible to provide a safe and energy efficient maneuver for the vehicle. The proceeding research is a completion of previous work [26-28] that takes a full consideration of details in task management and generalizes the applicability of the motion planners by realistic modeling of various underwater situations, which have not been fully addressed in previous papers.

All the paths between two sequential waypoints are modeled as continuous curves, which are parameterized by groups of several points using B-Spline algorithm. Thus, a mission can be modeled as a group of points (a sequence waypoints and sets of B-Spline control points). By taking all the points and associated conditions into account, creating and modifying a mission can be performed by defining and changing in the sequence and set of the points. This definition of mission makes it easy to be implemented using some computational optimization algorithm. This characteristic of the proposed idea is advantage to onboard software development in which suitability of the selected algorithms depends on the complexity and size of the problem.

In addition to the advantage of the proposed idea, this research takes the advantages of a specific composition Differential Evolution and Firefly Optimization Algorithms (DEFO) to fulfil the objectives of this research toward solving the stated problems associated with previous approaches, where DE is employed for the waypoint sequence generation and mission time management, and Firefly optimization algorithm applied for path generation between waypoints. The rest of layers are glue functions between these two layers. The DE is well-suited for dealing with complexity of task prioritization problem due to its discrete nature [23, 39]. On the other hand, the Firefly algorithm is advantaged to use an automatic subdivision approach that makes it specifically suitable and flexible in dealing with continuous problems (e.g. path planning) and multi-objective nonlinear problems [34-36]. Moreover, its control parameters can be tuned iteratively that increases convergence rate of the algorithm. To the best of the authors' knowledge, although many attempts have been carried out in the scope of vehicle routing-task allocation and trajectory/path planning for unmanned ground, aerial, surface and underwater vehicles, there is currently no particular research emphasized comprehensively in the scope of both motion planning and task scheduling approach in a systematic fashion, specifically the scope of underwater vehicles. Static current map and static uncertain obstacles along with real map data are taken into account for promoting the proposed path planner in handling real-world underwater situations. The path planner operates concurrently and back feed the environmental condition to higher level ask-allocate-route planner. A number of different approaches are tested in simulation and the stability and real-time applicability of the proposed method is shown.

The paper is structured in following sections. The mathematical modelling of the underwater operation terrain is provided by Section 2. Section 3 discusses about path planning problem taking kinodynamic of the AUV into account. Task allocation and routing is demonstrated and validated in Section 4. The discussion on validation of adaptive hierarchal model is provided by Section 5, and the Section 6 concludes the paper.

## 2   Mathematical Representation of the Waypoint Cluttered Ocean Terrain

For any of scientific, mine or military applications of the AUV, a sequence of tasks such as seabed habitat mapping, water sampling, mine detections, pipeline inspection, building subsea pipelines, seafloor mapping, payload delivery, surveillance, etc., is predefined and characterized in advance. The task sequence in this research is initialized with 15 different tasks and characterized as follows:

$$\aleph = \{\aleph_1,...,\aleph_{15}\}, \quad \forall \aleph_i, \quad \exists \rho_i, \xi_i, \delta_i \Rightarrow \begin{array}{l} \rho_i \sim U(1,10); \\ \xi_i \sim U(0,100); \\ \delta_i \sim U(20,200); \end{array} \qquad (1)$$

All parameters are denoted in "**Nomenclature**" table. The $U(a,b)$ represents a uniform distribution bounded to $(a,b)$ interval. Tasks for a specific mission are distributed in different areas of the operating field, in which placement of the tasks can be mapped and presented in a graph format where beginning and ending location of a task appointed with waypoints. Existence of a prior information about the terrain, location of starting and ending of each task (waypoints), and position of the global start/destination advances the AUV to accurately map the environment for the purpose of task-assign-routing and accurate path planning. Therefore, the terrain is mapped with an undirected weighted network denoted by $G = (P, E)$, where $P$ is the set of vertices in the graph that corresponds to waypoints and $E$ denotes the edges of the graph in which some of the edges are assigned with a specific task (presented by (2)).

$$G = (P, E) \Rightarrow \begin{array}{l} |P| = k; \\ |E| = m \end{array}; \quad \begin{array}{l} P(G): \{p^1,..., p^k\}; \\ E(G): \{e^1,..., e^m\}; \end{array} \Rightarrow e^{ij} = (p^i, p^j) \qquad (2)$$

The graph is promoted to be connected and the connections, which are assigned with a task, are weighted with a value more than one. The connection's weight is calculated based on attributes of the corresponding task (given by (1) and (3)).

$$\forall e^{ij} \ \exists \ w_{ij}, d_{ij}, t_{ij}$$

$$w_{ij} = \begin{cases} \dfrac{\rho_{ij}}{\xi_{ij}} & \text{if } e^{ij} \wedge \aleph_l \\ 1 & \text{otherwise} \end{cases} \quad (3)$$

$$d_{ij} = \sqrt{(p_x^j - p_x^i)^2 + (p_y^j - p_y^i)^2 + (p_z^j - p_z^i)^2}$$

$$t_{ij} = \dfrac{d_{ij}}{|\upsilon|} + \delta_{ij}$$

For modelling a realistic marine environment, a three dimensional terrain $\Gamma_{3D}:\{10\ km^2\ (x\text{-}y),\ 100\ m(z)\}$ is considered using Vincent gulf map (located in south Australia). The map is clustered to water zone (allowed for deployment) and coastal/uncertain areas presented by *Fig.1*. The clustered map is transformed to matrix format, in which the water-covered areas on the map filled with value of 1 and the coastal/ uncertain areas assigned with a value between [0, 0.3]. The waypoints (vertices of the network) are initialized in eligible sections for operation (water covered area) as follows:

$$\forall p^i \in P \ \Rightarrow \ \begin{matrix} p_{x,y}^i \sim U(0,10000) \\ p_z^i \sim U(0,100) \end{matrix} \ \Rightarrow \ p_{x,y,z}^i \in \{Map=1\} \quad (4)$$

here, $\{Map=1\}$ denotes the water covered area. In *Fig.1*, the subsections of coastal area, uncertain risky area, and water covered area are presented with black, grey and white colors, respectively. The area of 10 $km^2$ is used for task assign-routing approach and a sub-area of 3.5 $km^2$ (presented by red square in *Fig.1*) is selected for testing the performance of the local path planner. Additional to offline map, terrain is randomly covered by uncertain obstacles, in which their coordinates can be measured by the sonar sensors with a specific uncertainty modelled with a Gaussian distributions and presented in a circular format radiating out from the center of the object. Each obstacle $\Theta$ is characterised by its position bounded to position of $p^a$ and $p^b$ waypoints ($\Theta_p \sim \mathcal{N}(0,\sigma_p^2) \in [p^a_{x,y,z}, p^b_{x,y,z}]$), radius $\Theta_r \sim \mathcal{N}(0,\ \sigma_r^2)$ and uncertainty $\Theta_{Ur} \sim \mathcal{N}(\Theta_p,\sigma_0=\Theta_r)$, where the value of $\Theta_r$ in each iteration $t$ is independent of its previous value.

On the other hand, current is an important parameter that can affect vehicle's motion along the generated trajectory. The water current map can be captured from remote observations provided by satellite or from numerical estimation models. Different type of predictive ocean current models have been constructed previously [21, 29-31]. In this research, information of static 2D turbulent current has been employed, as the deep ocean current fields do not change immediately. The current dynamics is estimated and modelled using superposition of multiple Lamb vortices and 2-D Navier-Stokes equation [32]. The physical model employed by the AUV to diagnose the current velocity field is mathematically described by:

$$V_C = (u_c, v_c) \Rightarrow \begin{cases} u_c(\vec{S}) = -\Im \dfrac{y - y_0}{2\pi(\vec{S} - \vec{S}^O)^2}\left[1 - e^{\dfrac{-(\vec{S} - \vec{S}^O)^2}{\ell^2}}\right] \\ v_c(\vec{S}) = \Im \dfrac{x - x_0}{2\pi(\vec{S} - \vec{S}^O)^2}\left[1 - e^{\dfrac{-(\vec{S} - \vec{S}^O)^2}{\ell^2}}\right] \end{cases} \quad (5)$$

where, $S=(x,y)$ represents a 2-D space, $S^o$ denotes the center of the vortex, $\ell$ is the radius of the vortex, and $\Im$ is the strength of the vortex (tourbillon). Based on the tuning parameters such as the center, radius, and strength of the vortex (tourbillon) and rough knowledge of littoral water behavior, the equation can represent an acceptable current dynamic behavior.

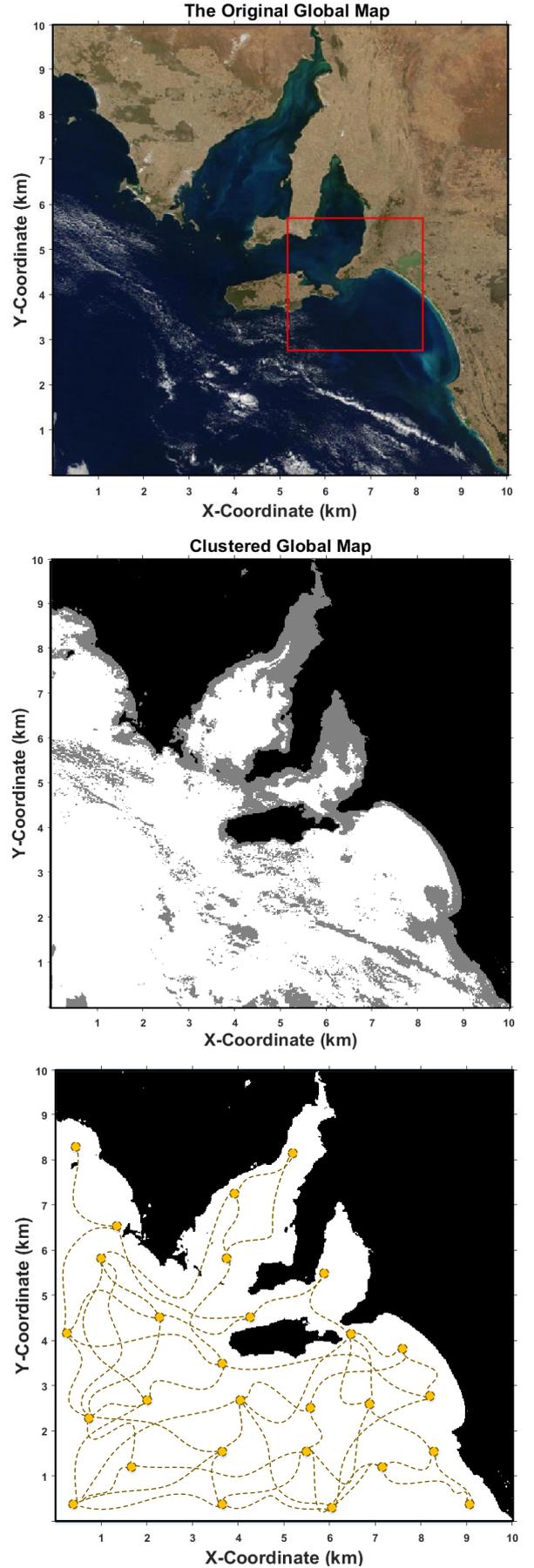

**Fig.1.** The original and clustered map of the Vincent gulf and graph representation of the operation network covered by waypoints.

## 3 Structure of the FO-Based Time-Optimal Path Planner

The path planning is an NP-hard optimization problem in which the main goal is to minimize the path length, avoid crossing collision borders, and coping current variations over the time. Adverse current can push the vehicle to an undesired direction and lead extra battery consumption, while a desirable current can motivate its motion and cause save to energy consumption. AUV's dynamics and kinematic are described using a set of ordinary differential equations given by (6) and (7) [33]. *Figure 2* shows the vehicle's state variables of body frame $\{b\}$ and NED (North-East-Depth) $\{n\}$-frame that provides its motion with six degree of freedom. The potential trajectory $\wp_i$ in this research is generated based on B-Spline curves captured from a set of control points $\vartheta:\{\vartheta^1_{x,y,z},...,\vartheta^i_{x,y,z},...,\vartheta^n_{x,y,z}\}$, defined by (8).

$$\eta:(X,Y,Z,\varphi,\theta,\psi); \quad \upsilon:(u,v,w,p,q,r) \tag{6}$$

$$\begin{bmatrix}\dot{X}\\\dot{Y}\\\dot{Z}\end{bmatrix}=\begin{bmatrix}^n_bR\end{bmatrix}\begin{bmatrix}u\\v\\w\end{bmatrix}; \quad \begin{bmatrix}^n_bR\end{bmatrix}=\begin{bmatrix}\cos\psi\cos\theta & -\sin\psi & \cos\psi\sin\theta\\ \sin\psi\cos\theta & \cos\psi & \sin\psi\sin\theta\\ -\sin\theta & 0 & \cos\theta\end{bmatrix} \tag{7}$$

$$\begin{cases}X=\sum_{i=1}^n \vartheta_{x(i)}B_{i,K}\\ Y=\sum_{i=1}^n \vartheta_{y(i)}B_{i,K}\\ Z=\sum_{i=1}^n \vartheta_{z(i)}B_{i,K}\end{cases} \tag{8}$$

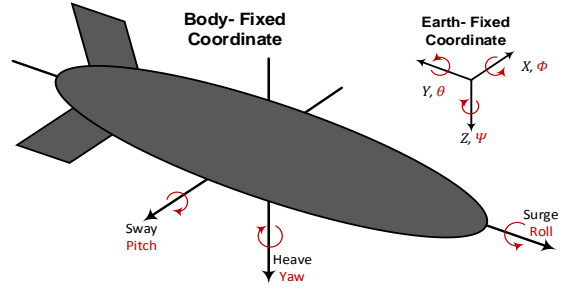

Fig.2. Vehicle's coordinates in NED and Body frame accordingly

here, the $\eta$ is state of the vehicle in $\{n\}$-frame; $X,Y,Z$ give vehicle's position along the generated path, and $\varphi,\theta,\psi$ are the Euler angles of roll, pitch, and yaw, respectively. The $\upsilon$ is AUV's velocity in the $\{b\}$-frame; $u,v,w$ are AUV's directional velocities of surge, sway and heave; and $p,q,r$ are the AUV's rotational velocities in the x-y-z axis. The $[^n_bR]$ is a rotation matrix that transforms the body frame $\{b\}$ into the NED frame $\{n\}$. The $B_{i,K}$ is the curve's blending functions, and $K$ adjusts smoothness of the curve. Water currents continually affect the vehicle's motion, so the vehicle's angular velocity components along the path curve $\wp$ is calculated considering water current correlation:

$$\theta=\tan^{-1}\left(-\left|\vartheta_{z(i+1)}-\vartheta_{z(i)}\right|\bigg/\sqrt{\left(\vartheta_{x(i+1)}-\vartheta_{x(i)}\right)^2+\left(\vartheta_{y(i+1)}-\vartheta_{y(i)}\right)^2}\right) \tag{9}$$
$$\psi=\tan^{-1}\left(\left|\vartheta_{y(i+1)}-\vartheta_{y(i)}\right|\bigg/\left|\vartheta_{x(i+1)}-\vartheta_{x(i)}\right|\right)$$

$$\begin{cases}u_c=|V_C|\cos\theta_c\cos\psi_c\\ v_c=|V_C|\cos\theta_c\sin\psi_c\end{cases} \Rightarrow \begin{matrix}u=|\upsilon|\cos\theta\cos\psi+|V_C|\cos\theta_c\cos\psi_c\\ v=|\upsilon|\cos\theta\sin\psi+|V_C|\cos\theta_c\sin\psi_c\\ w=|\upsilon|\sin\theta\end{matrix} \tag{10}$$

$$\wp=[X,Y,Z,\psi,\theta,u,v,w] \tag{11}$$

Control points should be located in respective search region constraint to predefined upper and lower bounds of $\vartheta_i \in [U_\vartheta, L_\vartheta]$ in Cartesian coordinates, where the $L_\vartheta$ is the lower bound that corresponds to location of the start point ($L_\vartheta \equiv p^a_{x,y,z}$) and $U_\vartheta$ is the upper bound that corresponds to location of the target point ($U_\vartheta \equiv p^b_{x,y,z}$).

### 3.1 FO Algorithm on Path Planning Approach

To generate trajectory by B-Spline curves, the FO algorithm is adapted to accurately locate the control points ($\vartheta$) of a candidate curve ($\wp$) in the solution space between the predefined upper bound ($U^i_\vartheta$) and lower bound ($L^i_\vartheta$). Appropriate adjustment of control points play a substantial role in determining the optimal path. A firefly in this context corresponds to a candidate solution (path) involving a distinct number of control points. FO is another swarm-intelligence-based meta-heuristic algorithm inspired from the flashing patterns of fireflies, in which the fireflies attracted to each other based on their brightness [34, 35]. As the distance of fireflies increases their brightness gets dimmed. The less bright firefly approaches to the brighter one. The firefly's brightness is determined through the perspective of the objective function. Attraction of each firefly is proportional to its brightness intensity received by adjacent fireflies and their distance $L$. The attraction factor $\beta$, their distance $L$ and movement of a firefly $i$ toward the brighter firefly $j$ is calculated by

```
FO based Path Planner
Initialization phase:
▪ Initialize population of fireflies χ_q (q =1,2,…,n) with the control points ℘_i
▪ Define light absorption coefficient ε
▪ Initialize the attraction coefficient β_0
▪ Set the damping factor of κ
▪ Initialize the randomness scaling factor of α_0
▪ Set the parameter of randomization α_t
▪ Set the maximum iteration MaxIter
▪ Set the number of population nPop
  For t =1: MaxIter
    For i =1: nPop
      Reconstruct a path according to χ_i
      Evaluate the path C_℘(χ_i(t))
      Update light intensity of χ_i
      For j =1:i
        Reconstruct a path according to χ_j
        Evaluate the path C_℘(χ_j(t))
        Update light intensity of χ_j
        If (β_j > β_i),
          Move firefly i towards j
        end
      end
    end
    Rank the fireflies and find the current best
  end
Output result
```

Fig.3. Pseudocode of FO mechanism on path planning approach.

$$L_{ij} = \|\chi_j - \chi_i\| = \sqrt{\sum_{q=1}^{d}(x_{i,q} - x_{j,q})^2}$$

$$\beta = \beta_0 e^{-\varepsilon L^2}$$

$$\chi_i^{t+1} = \chi_i^t + \beta_0 e^{-\varepsilon L_{ij}^2}(\chi_j^t - \chi_i^t) + \alpha_t \varsigma_i^t \quad (12)$$

$$\alpha_t = \alpha_0 \kappa^t, \quad \kappa \in (0,1)$$

where, $x_{i,q}$ is the $q^{th}$ component of the firefly $\chi_i$ coordinate in $d$ dimensions; $\beta_0$ is the attraction value at $L=0$, $\alpha_t$ is the randomization parameter that control the randomness of the movement and can be tuned iteratively. The $\alpha_0$ is the initial randomness scaling value and $\kappa$ is a damping factor. The $\varsigma_i^t$ is a random vector generated by a Gaussian distribution at time $t$. There should be a proper balance between engaged parameters, because if the $\beta_0$ approaches to zero, the movement turns to a simple random walk, while $\varepsilon = 0$ turns it to a variant of particle swarm optimization [34]. The pseudocode of FO process on path planning approach is provided by *Fig.3*. The candidate path solutions are coded by fireflies and tend to be optimized iteratively toward the best solution in the search space according to given optimization criterion.

The FO algorithm is advantaged uses an automatic subdivision approach that makes it more efficient comparing to other optimization algorithms. Such an automatic subdivision nature increases convergence rate of the algorithm, motivates fireflies to find all optima iteratively and simultaneously, which makes the FO specifically suitable and flexible in dealing with continuous problems, highly nonlinear problems, and multi-objective problems [36]. The control parameters in FO can be tuned iteratively that is another reason for its fast increases convergence.

### 3.2 Path Optimization Criterion

The AUV is considered to have constant thrust power; therefore, the battery usage for a path is a constant function of the distance travelled. Performance of the generated trajectory is evaluated based on overall collision avoidance capability and length of the path. The resultant path should be safe and feasible. The environmental constraints are associated with the depth limitation for vehicles deployment, forbidden zones of map or intersecting any obstacle, and coping current flow that may causes drift between desired and actual deployment of the vehicle. The water current causes drift between desired and actual deployment of the vehicle. AUV's surge-sway velocities and its yaw-pitch orientation should be constrained to $u_{max}$, $[v_{min}, v_{max}]$, $\theta_{max}$, and $[\psi_{min}, \psi_{max}]$ in all states along the path. The path cost is formulated as follows:

$$\forall \wp_{x,y,z}^j \to L_\wp = \sum_{i=p_{x,y,z}^a}^{|\wp|} \sqrt{(X_{i+1}(t)-X_i(t))^2 + (Y_{i+1}(t)-Y_i(t))^2 + (Z_{i+1}(t)-Z_i(t))^2} \quad (13)$$

$$\nabla_{\Sigma_{M,\Theta}} = \begin{cases} 1 & \wp_{x,y,z}(t) = Coast: Map(x,y) = 1 \\ 1 & \wp_{x,y,z}(t) \cap \bigcup_{N\Theta} \Theta(\Theta_p, \Theta_r, \Theta_{Ur}) \\ 0 & Otherwise \end{cases}$$

$$\nabla_\wp = \begin{cases} \varepsilon_{zmin} \times \min(0; Z(t) - Z_{min}) \\ \varepsilon_{zmax} \times \max(0; Z(t) - Z_{max}) \\ \varepsilon_u \times \max(0; u(t) - u_{max}) \\ \varepsilon_v \times \max(0; |v(t)| - v_{max}) \\ \varepsilon_\theta \times \max(0; \theta(t) - \theta_{max}) \\ \varepsilon_\psi \times \max(0; |\dot\psi(t)| - \psi_{max}) \\ \varepsilon_{\Sigma_{M,\Theta}} \times \nabla_{\Sigma_{M,\Theta}} \end{cases} \quad (14)$$

$$C_\wp = L_\wp + \sum_{i=1}^{n} Q_i f(\nabla_\wp)$$

here, the $Q_i f(\nabla_\wp)$ is a weighted violation function that respects the AUV kinodynamic and collision constraints including depth violation ($Z$) to prevent the path from deviating outside the vertical operating borders, surge ($u$), sway ($v$), yaw ($\psi$), pitch ($\theta$) violations, and the collision violation ($\nabla_{\Sigma_{M,\Theta}}$) specified to prevent the path from collision danger. The $\varepsilon_{zmin}$, $\varepsilon_{zmax}$, $\varepsilon_u$, $\varepsilon_v$, $\varepsilon_\theta$, $\varepsilon_\psi$, and $\varepsilon_{\Sigma_{M,\Theta}}$ respectively denote the impact of each constraint violation in calculation of total path cost $C_\wp$.

### 3.3 Evaluation of the FO-based Local Path Planner

To evaluate the performance of the local path planner in this research, real map data, uncertain no-flying zones, uncertain static obstacles, and static water current map are considered to cover different possibilities of the real world situations. The vehicle moves with constant water-referenced velocity $\upsilon$. As regards, current velocity is proportional to the cube root of the thrust, similarly, the AUV considered to have constant thrust power, and therefore, the battery usage for a path is a constant multiple of the distance travelled. Thus, the goal of the path planner is to take the shortest battery efficient trajectory between two waypoints, in which the trajectory avoids colliding forbidden zones, handles undesired current flows and take use of desirable current to save more battery. Assumptions play important role for the controller in coping the current disturbance and to accurately drive the AUV along the planned trajectory. The vehicle control and guidance strategies along with full dynamics of the system has been investigated previously [37, 38]. Current research takes the advantages of the previous investigations to implement the path generation module. The B-spline paths' curvature is obtainable by the vehicle's radial acceleration and angular velocity constraints. The FO parameters are configured as follows: number of B-Spline control points is set on 5. The Fireflies population is set on 100, the initial attraction coefficient $\beta_0$ is set on 2, and light absorption coefficient $\varepsilon$ is assigned with 1. The damping factor of $\kappa$ is assigned with $\kappa \in [0.95, 0.97]$. The scaling variations is defined based on initial randomness scaling factor of $\alpha_0$. The parameter of randomization is set on 0.4.

*Figure 4* represents the path behavior in different situations mentioned above where the complexity of the operating window increase from *Fig.4(a)* to *Fig.4(c)*. The current field computed from a random distribution of 12 Lamb vortices in 100×100 grid; hence, each pixel in *Fig.4(a,b)* contains a current arrow and corresponds to area of 35 $m^2$. The position of start and target points presented by, red and green squares, respectively. In *Fig.4* a set of pareto-optimum paths are generated in accordance with terrain situation in which the best generated path is proposed by thicker line and the sub optimum trajectories are presented by thinner lines. *Figure 4(a)* shows the paths behaviour to current flow. Afterward, the terrain modelled to be more complex and the path efficiency of collision avoidance is investigated encountering static current information and uncertain static obstacles (no-flying zones) in which their uncertainty grows by time in a circular format (given in *Fig.4(b)*). The *Fig.4(c)* investigates the accuracy of the generated paths in recognizing forbidden coastal areas on the map while three different target points from the same starting point are considered for better representation of path behaviour in dealing with water vortexes. A k-means clustering is applied to clarify water covered and coastal areas. Considering the path deformation it is noteworthy to hint the efficiency of the proposed method in adapting current flows whether in avoiding undesirable turbulent or driving the compliant current arrows; more specifically, optimum path and also its alternatives are accurately adapt with current arrows and avoid colliding forbidden edges that is a critical concern in safe deployment.

On the other hand, the vehicular constraints and boundary conditions on AUV actuators and state also are taken into account for realistic modelling of the AUV operation. As discussed earlier the violation function for the path planners is defined as a combination of the vehicle's depth, surge, sway, theta, yaw and collision violations. According to (14), the boundaries for given constraints are defined as follows: the $z_{min}=0(m)$; $z_{max}=100(m)$; $u_{max}=2.7(m/s)$; $v_{min}=-0.5(m/s)$; $v_{max}=0.5(m/s)$; $\theta_{max}=20$ (*deg*); $\psi_{min}=-17$ (*deg*) and $\psi_{max}=17$ (*deg*). Total violation and cost variations of the path population over 100 iterations are presented by *Fig.6* and *Fig.5,* respectively. As shown by *Fig.5*, the generated trajectory performs a great fitness encountering all constraints. The cost variations shows that algorithm experiences a moderate convergence by passing iterations as the cost variation range decreases in each iteration. This means algorithm accurately converges to the optimum solution with minimum cost. It is further noted from *Fig.6*, the proposed algorithm accurately satisfies the proposed constraints as the variations of total violation for path population is diminishing iteratively, which means algorithm successfully manages the path toward eliminating all defined violation factors.

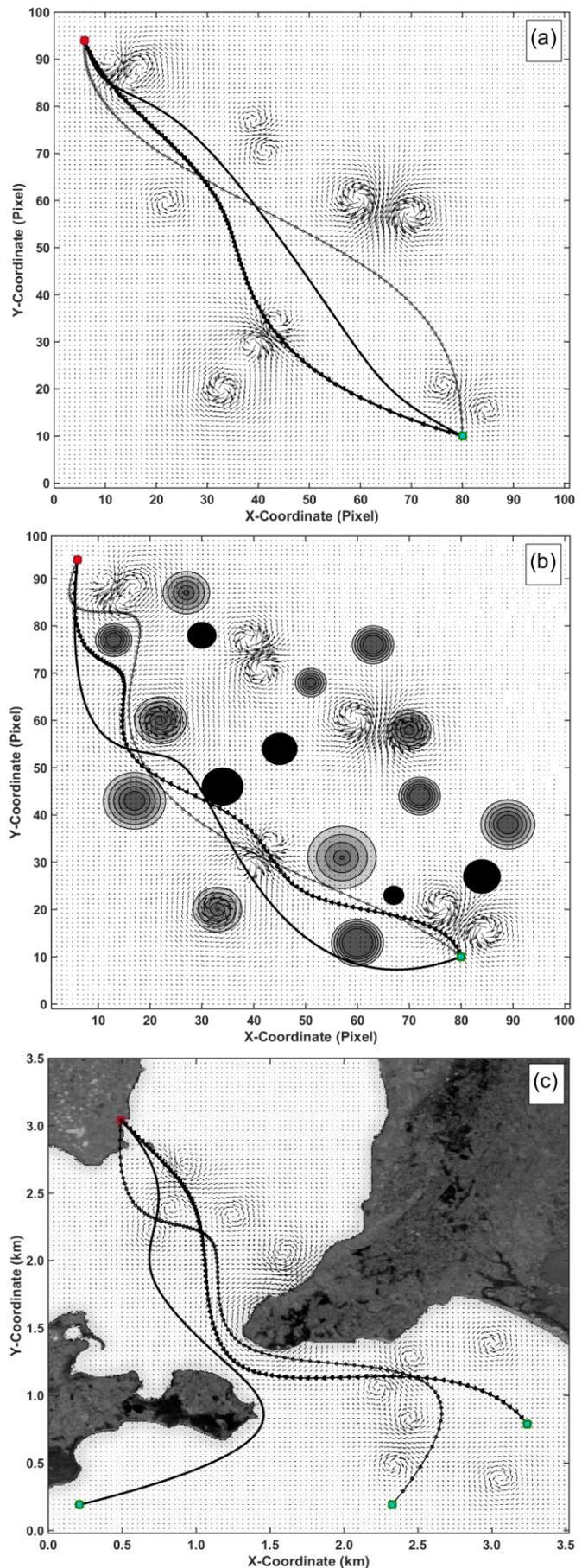

**Fig.4.** **(a)** Path adaption to current arrows in a static current map; **(b)** Path behavior in avoiding colliding uncertain forbidden zones and obstacles in presence of current flow; **(c)** Paths behavior in recognizing forbidden coastal areas and adapting current flows.

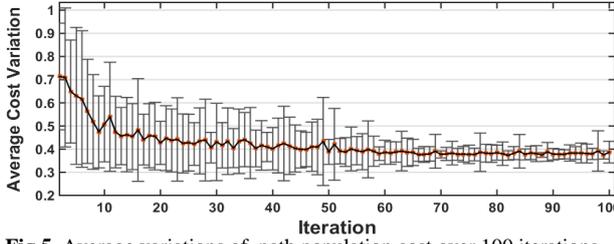 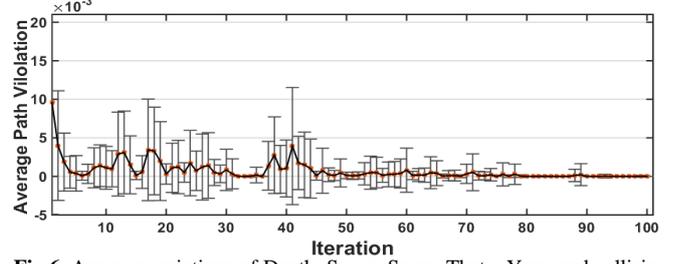

**Fig.5.** Average variations of path population cost over 100 iterations.   **Fig.6.** Average variations of Depth, Surge, Sway, Theta, Yaw, and collision violation corresponding to generated path population over 100 iterations.

The shortcomings with the path planner appears when the vehicle is required to operate in a large scale terrain, as it should compute a large amount of data repeatedly and estimate dynamicity of the terrain adaptively. Moreover, path planner only deals with vehicles guidance from one point to another and do not deal with mission scenario or task assignment considerations. To address the mission scenario and tasks priority assignment and also to handle the shortcomings of the local path planning, the graph route planner operates in a higher level to give a general overview of the terrain and cut off the operating area to beneficial zones for vehicles deployment in the feature of the global route including a sequence of tasks.

## 4   Mathematical Representation of the Routing Problem

In a terrain that covered by several waypoints, the vehicle is requested to furnish maximum number of highest priority tasks with minimum risk percentage in the total available time. With respect to graph-like terrain modelled in section 2, the route planner tends to find the best fitted route to the total available time, involving the best sequence of waypoints in which the total collected weight by the route is maximized, which means the edges that containing the highest priority tasks are selected and ordered in a manner to guide the vehicle toward its final destination. On the other hand, on-time termination of the mission should be guaranteed which means the route travel time should not exceed the total available time that is started to counting inversely from the beginning of the mission. Hence, the problem is a restricted multi-objective optimization problem very similar to combination of the TSP and Knapsack problem. In the preceding discussion, the mathematical representation of the route planning problem for AUV in $\Gamma_{3D}$ terrain is describes as follows:

$$\Re_k = \left( p^S_{x,y,z}, \ldots, p^i_{x,y,z}, \ldots, p^D_{x,y,z} \right)$$
$$\forall \left( p^i_{x,y,z}, p^j_{x,y,z} \right) \; \exists \; w_{ij}, d_{ij}, t_{ij} \Rightarrow \begin{cases} w_{ij} > 1 & if \; \exists \; \aleph_{ij} \\ w_{ij} = 1 & f \; \exists! \; \aleph_{ij} \end{cases} \qquad (15)$$

Here, the $\Re$ denotes a route that is started from node $p^s$ and ended at $p^D$, in which any edge between two arbitrary points of $(p^i, p^j)$ involves a weight ($w_{ij}$) that represents the value of the selected edge (its corresponding distance ($d_{ij}$) may assigned with a task $\aleph$ or not). To address the mentioned above requirements, the route time $T_\Re$ should approach the total time $T_\tau$ and the captured weight by the route should be maximized as formulated by (16) and (17).

$$T_\Re = \sum_{\substack{i=0 \\ j \neq i}}^{n} s_{eij} \times t_{ij} = \sum_{\substack{i=0 \\ j \neq i}}^{n} s_{eij} \times \left( d_{ij} / |v| + \delta_{\aleph ij} \right), \quad s_{eij} \in \{0,1\} \qquad (16)$$

$$\max \left( \sum_{\substack{i=0 \\ j \neq i}}^{n} s_{eij} \times w_{ij} \right) \quad \& \quad \begin{matrix} \min(|T_\Re - T_\tau|) \\ s.t. \\ \max(T_\Re) < T_\tau \end{matrix} \qquad (17)$$

where, $s_{eij}$ is the selection variable that gets 1 for the selected edges and 0 for the rest. The planned route should be applicable and logically feasible according to feasibility criteria's given below.

- The route must be commenced and ended with index of the specified start and destination nodes.
- The route must exclude the non-existent edges in the graph.
- The multiple appearance of a node in a route makes it inefficient by wasting time on repeating a task.
- The route must pass an edge maximum for once, hence the visited edges should be eliminated.

To produce a feasible route a priority vector and the graph adjacency matrix is employed. This process is carried out in first step of the DE-based routing approach (initialization phase).

## 4.1 Differential Evolution on Constraint Optimization Problem of Task Assign-Route Planning

The Differential Evolution (DE) algorithm [39] is an improved version of genetic algorithm that uses similar operators of selection, mutation, and crossover. The DE constructs better solutions and faster optimization due to use of real coding of floating point numbers in presenting problem parameters. The algorithm has a simple structure and mostly relies on differential mutation operation and non-uniform crossover as a search mechanism. Then applies selection operator to converge the solutions toward the desirable regions in the search space. The offspring's produced by crossover or mutation operators are inherited in unequal proportions from the previous solution vectors. In an optimization problem a cost function is required to be minimized. The process of the DE algorithm is clarified in the following subsections.

i. **Initialization:** In first step, the initial population of the solution vectors $\chi_i$, ($i=1,...,nPop$) is initialized using a randomly generated uniform probability vector and graph adjacency matrix in order to keep the solutions feasible and restricted to valid search space (respecting to routing feasibility criteria's defined above). For this purpose, nodes are selected and added to the potential route sequence based on their corresponding value in priority vector and adjacency matrix. To prevent repeated visit to a node in a route, the corresponding priority value of the selected node gets a big negative value; then, the adjacency matrix gets updated by eliminating the visited edges. In fact, using the adjacency matrix at this stage prevents appearance of non-existent edges in the graph. For more detail refer to [24]. In a case that the route is terminated with a non-destination node, the index of the last node in the sequence gets replaced by index of the destination node. Feasibility of the solutions is checked iteratively. Applying the evolution operators improves the solution space iteratively.

ii. **Mutation:** The effectual modification of the mutation scheme is the main idea behind impressive performance of the DE algorithm, in which a weighted difference vector between two population members to a third one is added to mutation process that is called *donor*. Three different individuals of $\chi_{r1,G}$, $\chi_{r2,G}$ and $\chi_{r3,G}$ are selected randomly from the same generation $G$, which one of this triplet is randomly selected as the *donor*. So, the mutant solution vector is produced by

$$\chi_{i,G} \begin{cases} i=1,...,nPop \\ G=1,...,G_{\max} \end{cases}$$
$$\dot{\chi}_{i,G} = \chi_{r3,G} + F_s(\chi_{r1,G} - \chi_{r2,G}) \qquad (18)$$
$$r1, r2, r3 \in \{1,...,nPop\},$$
$$r1 \neq r2 \neq r2 \neq i, \quad F_s \in [0,1+]$$

where, $nPop$ is the number of routes (solutions) population in DE, $G_{max}$ is the maximum number of iterations (generations). The $F_s$ is a scaling factor that controls the amplification of the difference vector ($\chi_{r1,G} - \chi_{r2,G}$). Giving higher value to $F_s$ promotes the exploration capability of the algorithm. The proper donor accelerates convergence rate. In this approach, the *donor* is determined randomly with uniform distribution as follows:

$$donor = \sum_{i=1}^{3} \left( \lambda_i \Big/ \sum_{j=1}^{3} \lambda_j \right) \chi_{ri,G}, \qquad (19)$$

where, $\lambda_j \in [0,1]$ is a uniformly distributed value. The mutant individual $\dot{\chi}_{i,G}$ and parent individual $\chi_{i,G}$ are then shifted to the crossover (*Recombination*) operation.

iii. **Crossover:** The parent vector to this operator is a mixture of individual $\chi_{i,G}$ from the initial population and the mutant individual $\dot{\chi}_{i,G}$. The produced offspring $\ddot{\chi}_{i,G}$ from the crossover is described by

$$\begin{cases} \chi_{i,G} = (x_{1,i,G},...,x_{n,i,G}) \\ \dot{\chi}_{i,G} = (\dot{x}_{1,i,G},...,\dot{x}_{n,i,G}) \Rightarrow \ddot{x}_{j,i,G} = \begin{cases} \dot{x}_{j,i,G} & rand_j \leq r_C \vee j=k \\ x_{j,i,G} & rand_j \leq r_C \wedge j \neq k \end{cases} \\ \ddot{\chi}_{i,G} = (\ddot{x}_{1,i,G},...,\ddot{x}_{n,i,G}) \quad j=1,...,n; \quad n \in [1,nPop] \end{cases} \qquad (20)$$

where, $k \in \{1,...,nPop\}$ is a random index chosen once for all population $nPop$. The second DE control parameter is the crossover ratio $r_C \in [0,1]$ that is set by user.

iv. **Evaluation and Selection:** The offspring produced by the crossover and mutation operations is evaluated according feasibility criteria, then the feasible solutions turn to cost evaluation and infeasible solutions returned back to the corresponding mutation or crossover operation. The cost function is defined in the next part (route optimization criterion). The best fitted solutions produced by evolution operators are selected and transferred to the next generation ($G+1$).

$$\dot{\chi}_{i,G+1} = \begin{cases} \dot{\chi}_{i,G} & C_\Re(\dot{\chi}_{i,G}) \leq C_\Re(\chi_{i,G}) \\ \chi_{i,G} & C_\Re(\dot{\chi}_{i,G}) > C_\Re(\chi_{i,G}) \end{cases}, \quad \ddot{\chi}_{i,G+1} = \begin{cases} \ddot{\chi}_{i,G} & C_\Re(\ddot{\chi}_{i,G}) \leq C_\Re(\chi_{i,G}) \\ \chi_{i,G} & C_\Re(\ddot{\chi}_{i,G}) > C_\Re(\chi_{i,G}) \end{cases} \qquad (21)$$

The performance of the offspring and parents are compared for each operator according to route cost function of $C_\Re$ and the worst individuals eliminated from the population. The process of DE is presented through the *Fig.7*.

In *Fig.7*, the difference between two individuals {1,2} is added to a third individual {3}. The mutant individual {4} is sent to the crossover operator. The most fitted candidate from initial individuals {1,2,3}, mutated individual {4} and shuffled individual {5} is selected as proposal individual to the next generation. DE parameters for routing-task assignment problem is configured as follows: the population size is set on 100, lower and upper bound of scaling factor is set on 0.2 and 0.8, respectively, and the crossover ratio is set on $r_C$=0.2.

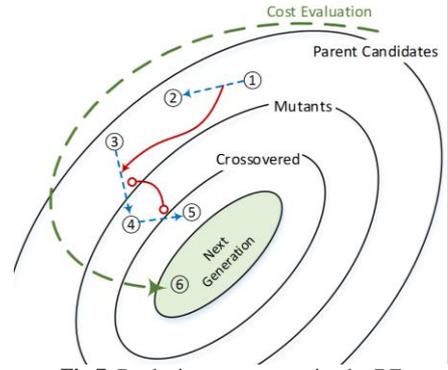

**Fig.7.** Producing new generation by DE

### 4.2 Route Evaluation Criterion

As mentioned earlier the local path planner operates in context of the graph route planner; hence, the path cost of $C_\wp$ has a direct impact on total cost of the global route because of its proportional relation to the travelled distance between pairs of waypoints ($d_{ij} \propto L_\wp$). On the other hand, the main goal is to maximize the weight of the selected edges in the graph, which means selecting the best sequence of highest priority tasks in a limited time. Therefore, the route cost of $C_\Re$ gets penalty when the $T_\Re$ for a particular route exceeds the $T_\tau$. Traversing the $L_\wp$ may take more time than what expected due to dynamic unexpected changes in the environment. The wasted time is compensated by carrying out a proper re-routing process. After visiting each waypoint in the global route, the re-planning criteria should be investigated (given in the next section). Hence; a computation cost encountered any time that re-planning is required. Thus, the route cost in the proceeding research is defined as follows:

$$C_\aleph = \sum_{\substack{i=0 \\ j \neq i}}^{n} s_{eij} \times \left( \frac{\xi_{\aleph,eij}}{\rho_{\aleph,eij}} \right), \quad s_{eij} \in \{0,1\} \tag{22}$$

$$C_\Re = \Phi_1 \left| \sum_{\substack{i=0 \\ j \neq i}}^{n} s_{eij} \times \left( \frac{C_{\wp ij}}{|\upsilon|} + \delta_{\aleph,\wp ij} \right) - T_\tau \right| + \Phi_2 C_\aleph + \sum_{1}^{r} T_{compute}$$

s.t.
$$\forall \Re \Rightarrow \max(T_\Re) < T_\tau \tag{23}$$

where $\Phi_1$ and $\Phi_2$ are two positive coefficients determine amount of participation of each factor in determination of the route cost, $T_{compute}$ is the time spent for checking the re-routing criteria (given below), and $r$ is repetition of the re-planning procedure in a mission.

### 4.3 Re-routing Criterion

Trade-off between managing the mission available time ($T_\tau$) and mission objectives should be adaptively carried out by the graph route planner. The $T_\wp$ is calculated at the end of the trajectory according to (25) and gets compared to expected time $T_{exp}$ for traversing the corresponding distance of $d_{ij}$, in which the $T_{exp} \equiv t_{ij}$ is determined from the $T_\Re$ given by (16). If the $T_\wp$ exceeds the $T_{exp}$, it means the local path planner spent extra time for coping any probable raised difficulty (e.g. collision avoidance or copping current disturbance). Obviously, a part of the available time $T_\tau$ is taken for this purpose and initial global route is turned to be invalid to the remained time; thus, re-routing would be necessary in this condition. For re-routing process the visited edges in previous route gets eliminated from the graph adjacency matrix (so the graph complexity is reduced and search space shrinks); the $T_\tau$ gets updated; and the existing waypoint is considered as new start point for both path/route planners. The, the route planner is recalled to generate new route according to graph and time updates. This process continues iteratively until the AUV reaches to the final destination (success) or it runs out of time/battery (failure). The proposed DEFO model proceeds as the flowchart in *Fig.8* declares.

$$\forall \wp_{x,y,z}^{ij} \Rightarrow T_\wp = \frac{|L_\wp|}{|\upsilon|} + \delta_{\aleph,e_{ij}} + \wp_{CPU} \tag{25}$$

Where, $\delta_{\aleph,eij}$ is the completion time of task assigned edge $e_{ij}$ and $\wp_{CPU}$ is the path computation time.

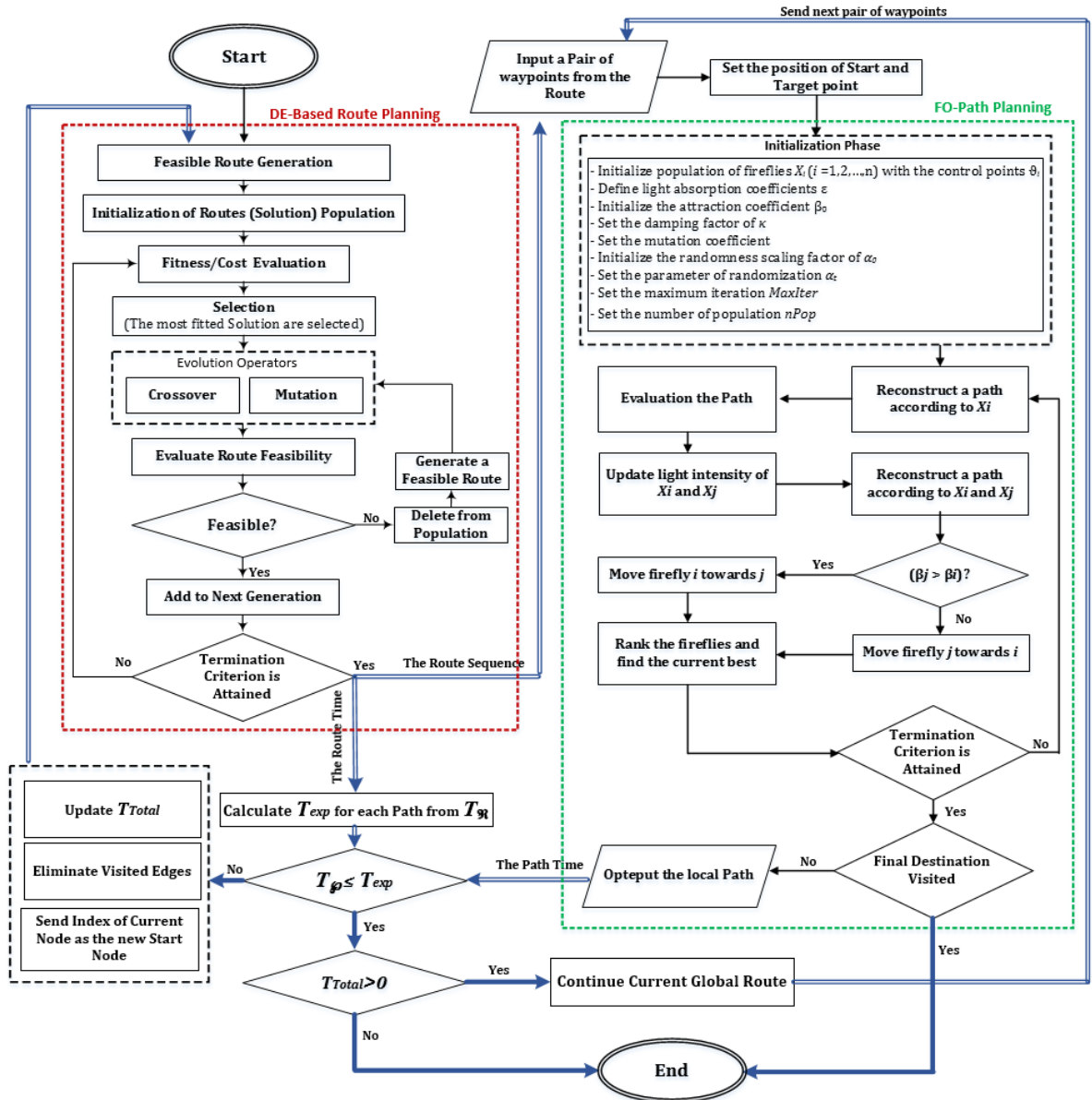

**Fig.8.** Flowchart of the proposed DEFO model including adaptive re-routing process

## 5   Performance Evaluation of the Adaptive Hierarchal DEFO Model

The graph route planner should be capable of generating a time efficient route with best sequence of tasks to ensure the AUV has a beneficial operation and reaches to the destination on-time as it's an obligatory requirement for vehicles safety. A beneficial operation is a mission that covers maximum possible number of tasks in a manner that total obtained weight by a route and its travel time is maximized but not exceed the time threshold. Along with an efficient route planning, the path planner in a smaller scale should be fast enough to rapidly react to prompt changes of the environment and generate an alternative trajectory that safely guides the vehicle through the specified waypoints in the optimum. Hence, the performance and stability of the model in satisfying the metrics of "obtained weight", "number of completed tasks", "cost of the route/path planners" and "total violation of the model" is investigated and analyzed in a quantitative manner through 50 Monte Carlo simulation runs and presented by *Fig.9*.

All mentioned performance indicators are investigated through 50 simulation of Monte Carlo. The Monte Carlo simulation performed in this section, is treated as a solid indicator in the state-of-the-art addressing to what extent the DEFO model can cope complexity (uncertainty) of the underlying mission scenarios. The Monte Carlo trials are initialized with realistic initial conditions that are analogous to real underwater mission scenarios. For all Monte Carlo runs, the quantity of waypoint is set to be changed with a uniform distribution between 30 to 50 nodes and the network topology also transforms randomly with a Gaussian distribution on the problem search space. The time threshold is set on $7.2 \times 10^3 (sec)$. A fixed set of 15 tasks, which characterized with risk percentage, priority value, and completion time, is specified and randomly assigned to some of the edges of the graph. The terrain is modelled as a realistic underwater environment, in which static ocean current map, real map data, and randomly generated uncertain static obstacles are considered in all experiments. Each experiment represents a mission.

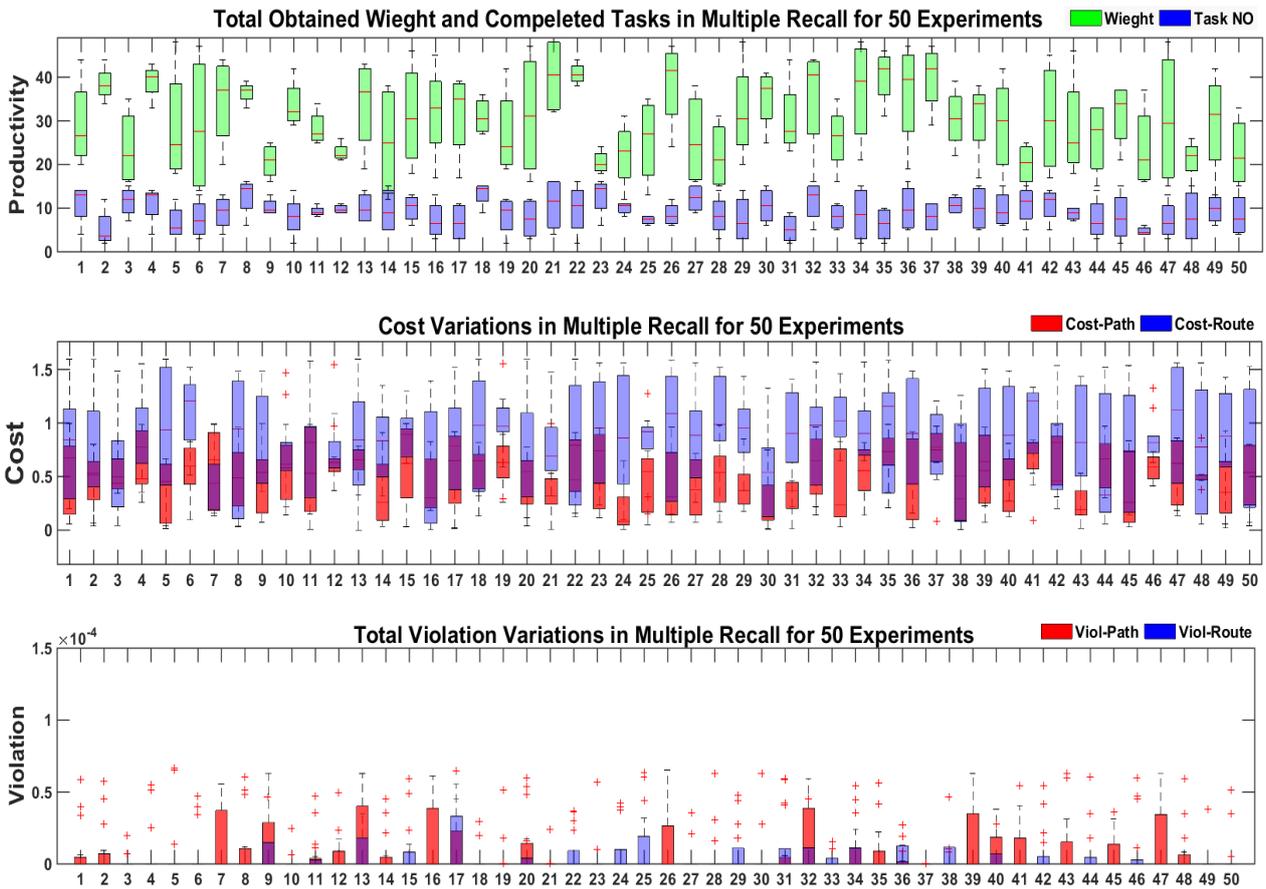

**Fig.9.** Statistical analysis of the model in terms of mission productivity according to total obtained weight and average number of completed tasks in a mission; cost and violation variation for both route/path planners in 50 Monte Carlo simulations.

Analysis of the result, captured from the Monte Caro simulations, indicates model's consistency and robustness against problem's space deformation (model complexity analysis). According to results presented in *Fig.9*, the cost variation for both planners shows the stability of the model in producing optimal solutions as the cost variation range stands in a specific interval for all experiments (mission). The path planner gets the violation if the generated path cross the collision edges, its depth drawn outside of the vertical operating borders or when the vehicles surge, sway, yaw and pitch parameters egress the defined boundaries; and the route planner gets violation if the route time $T_\Re$ exceeds the total available time for the mission $T_\tau$. Both path and route planners get recalled for several times in a particular mission. *Fig.9* shows that the model accurately satisfies all defined constraints as the violation value is considerably approached to zero in all 50 executions that is neglectable.

Another important performance indicator for such a combinatorial model is proper coordination of the higher and lower level motion planners in the system. To provide a precise concordance between two planners and real-time implementation of the model some other performance indexes should be highlighted additional to those which addressed above. Thus, one critical factor for both planners is having a short computational time to provide a concurrent synchronization. Fast operation of each planner keeps any of them from dropping behind the process of the other one, since appearance of such a delay damages concurrency of the entire system. Another significant performance metric that influences the synchronism of whole system is compatibility of the value of local path time ($T_\wp$) and the expected time ($T_{exp}$) in multiple operation of the local path planner. Therefore, the values of $T_\wp$ and $T_{exp}$ should be close to each other as it is critical for recognizing the requisition for re-routing. The models behavior in satisfying these two performance metrics also investigated in a quantitative feature indicated by *Fig.10*.

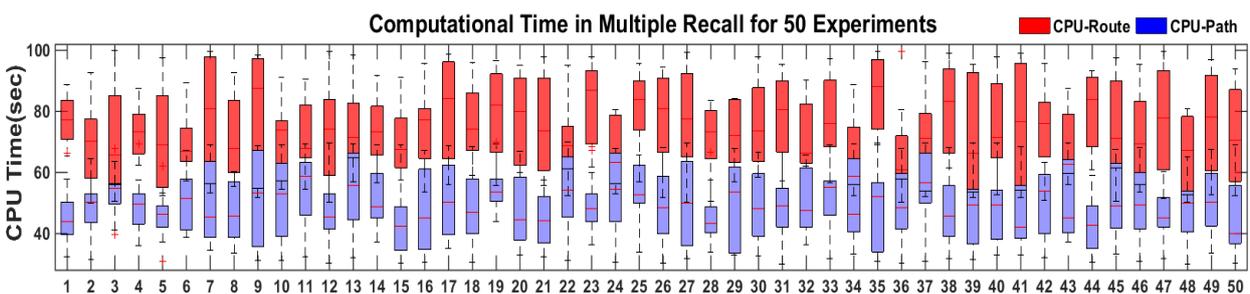

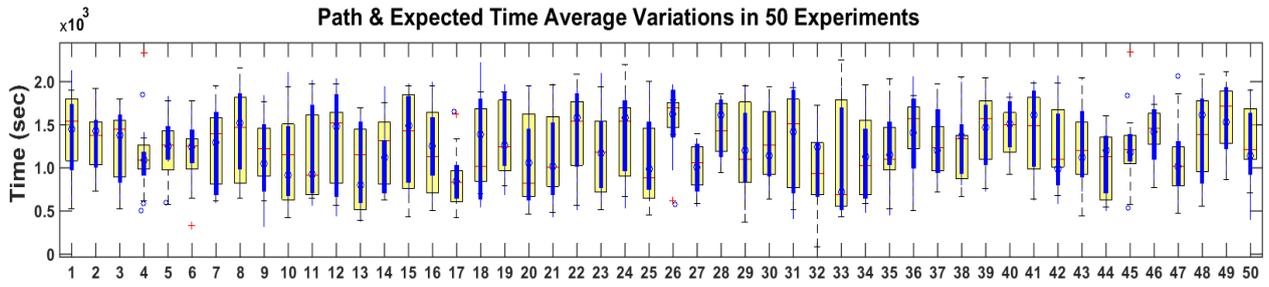

**Fig.10.** Real-time performance of the model, and compatibility of the value of $T_\wp$ (*presented by blue compact box plot*) and $T_{exp}$ (*presented by yellow transparent box plot*) in multiple operation of the local route planner through the 50 Monte Carlo simulation runs.

It is inferable from simulation result in *Fig.10* that variations of the CPU time for both Path/Route planners is drawn in a very narrow boundary in range of seconds in all executions that confirms real-time performance of this model in cooperating the changes of the environment and operation network. Considering the second plot in *Fig.10*, it is clear that the model accurately preserves the conformity and correlation between $T_\wp$ (*presented by blue compact box plot*) and $T_{exp}$ (*presented by yellow transparent box plot*) as their average variations for all executions are lied in similar range and very close to each other. This confirms efficient synchronization of the higher and lower level motion planners. Moreover, by considering the cost and violation variations in Fig.9 and the computational time (CPU time) in Figure 10, one can realize that the DEFO model is capable of finding feasible solutions with a fast convergence rate stemming from violations in order of $10^{-4}$ and CPU time in order of 100 second, respectively.

The best possible performance for the proposed model is completion of the mission with a minimum positive remaining time, which means the vehicle took maximum use of available time and terminated its mission before runs out of battery; thus, the most effective performance indicator for this model is its accuracy in mission time management. The result of simulation for 50 experiments is presented by *Fig.11*.

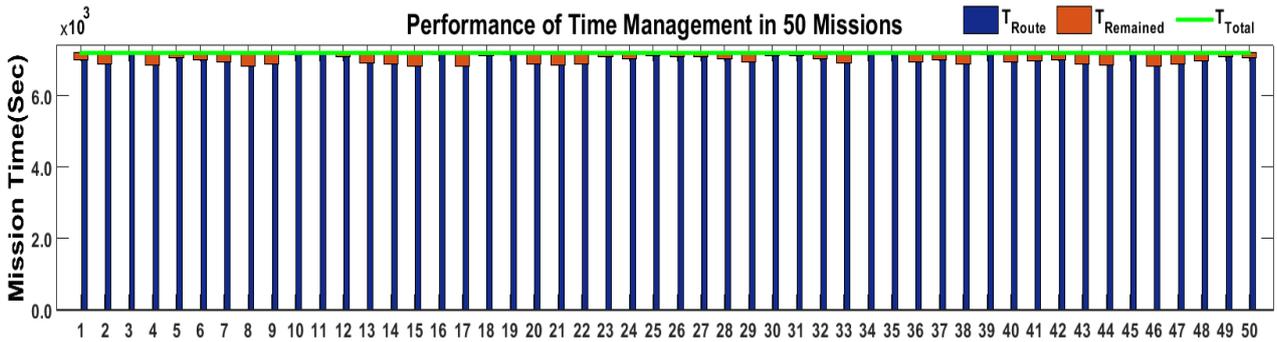

**Fig.11.** Statistical analysis of the model's performance in terms of mission time management and satisfying time constraint in 50 missions.

As declared by *Fig.11*, the remained time got a very small positive value in all missions that means no failure is occurred in this simulation, which is a remarkable achievement for having a confident and reliable mission as the failure is not acceptable for AUVs due to expensive maintenance in sever underwater environment. The time constraint is presented by green line in *Fig.11*. Obviously, the route time is maximized by minimizing the remaining time, which represents that how much of total available is used for completing different tasks in a single mission. Analyzing the variation of the route time and the remained time confirms supreme performance of the proposed novel model in mission reliability and excellent time management. For better understanding of the whole process, one experiment including three re-routing and 11 local path planning is shown by *Fig.12*.

Given a candidate initial route in a sequence of waypoints (edges assigned by tasks) along with environment information, the local path planner provides a trajectory to safely guide the vehicle through the waypoints in presence of ocean current and uncertain obstacles. In whole process, the remained time that initialized with the value equal to $T_\tau$, is counted inversely. Indeed, the remained time is the total available time that gets reduced by time. The local path planner incorporates any dynamic changes of the terrain while the vehicle is deploying between two waypoints; where in some cases its process may longer that cause the path drop behind the expected time for passing the corresponding distance; as occurred in passing the second edge in *Fig.12*. In such a case remained time gets updated by reducing the wasted time and the route planner is recalled to rearrange the sequences of tasks according to updated remained time, presented by the yellow thick line in *Fig.12*. As presented in *Fig.12*, number of 11 paths generated during 3 re-routing process in a single mission, where the discarded routes presented by dashed white lines. This synchronous process is frequently repeated until the AUV reach the destination that means mission success.

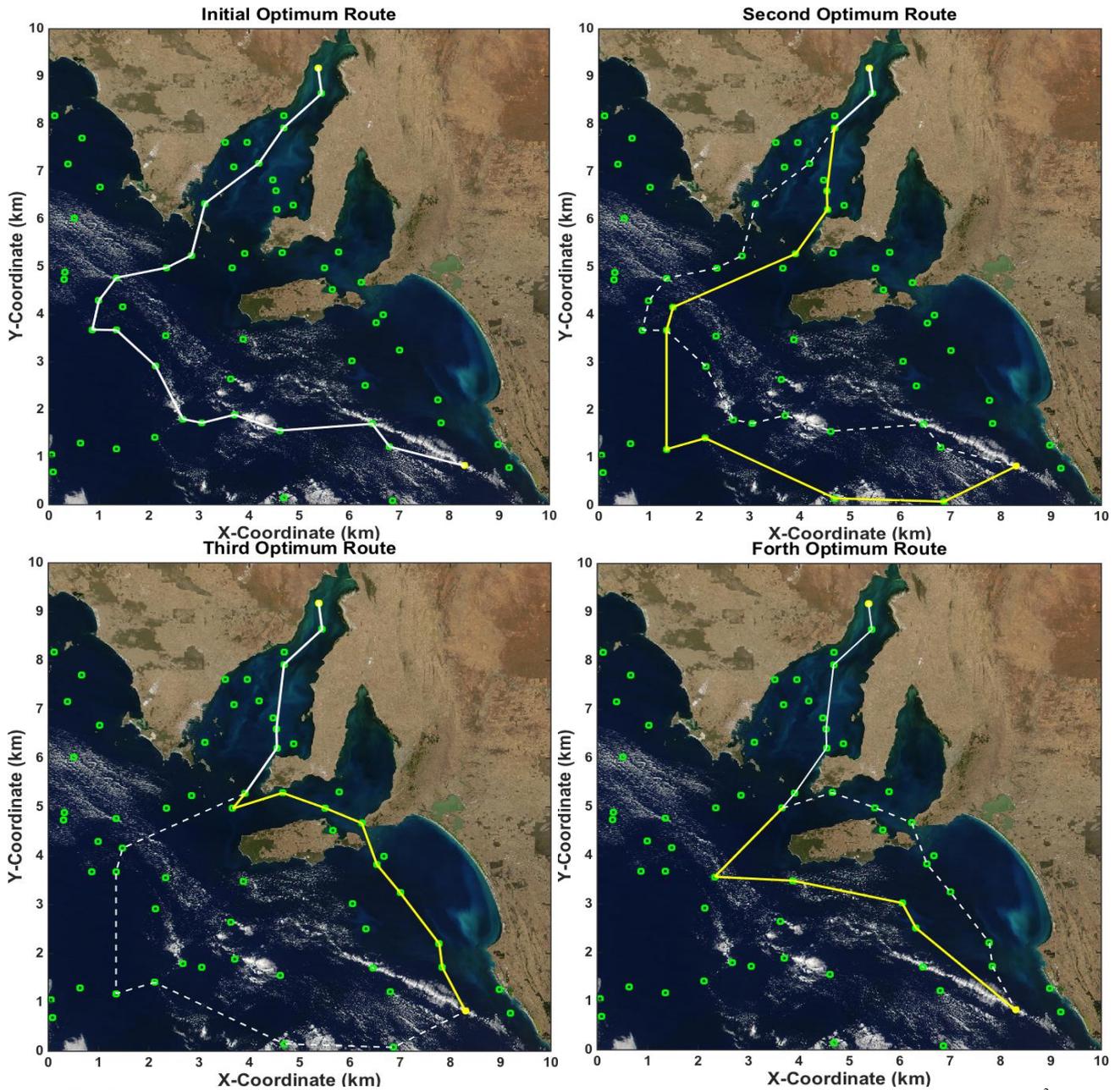

**Fig.12.** Process of route-path planning replanning and re-arrangement of order of edges (tasks) in a single mission in area of 10 $km^2$

The most important objective of this research is to validate performance of this model in efficient time management and guarantying on-time termination of the mission before vehicle runs out of time/battery as it is an important concern for mission success, which is analyzed and evaluated for 50 mission simulations in *Fig.11*. It is also noteworthy to mention from analysis of the Monte Carlo simulations in *Fig.9* and *Fig.10*, the variation ranges of performance metrics of total obtained weight, completed tasks, CPU time, total cost and violation is almost in a same range for all experiments that shows the stability and robustness of the model in dealing with environmental changes and random deformation of the graph topology.

## 6   Conclusion

This paper presented a hybrid strategy of task assign-route planning and path planning based on differential evolution and firefly optimization algorithms (called DEFO model) to maximize productivity of a single vehicle in a single mission within a limited time interval. Two higher/lower level motion planners are the cores of the proposed hybrid model where the route generator is responsible for prioritizing and managing the maximum number of tasks and the latter module generates an optimal-collision free path to govern the vehicles toward the goal of interest. For performance evaluation, the AUV operation was simulated in a three-dimension large terrain (almost 10 $km^2$ × 100 $m$), where static current map data along with uncertainty of the operation field is taken into account. The simulation results showed that the proposed model is efficient for increasing vehicle's autonomy of decision making in prioritizing tasks and mission time management. Providing this certain level of autonomy makes the vehicle and particularly path planner capable of using the favorable current flow for energy management, as well. Through the Monte Carlo trials it was inferred that the computational performance of the offered hybrid model is outstanding and works in a level of real-time performance. There was also a significant robustness with the model in terms of the terrain's variability and configuration changes of the allocated tasks. Future research will pay attention to detached modular architecture in which each layer of the architecture provides a specific level of autonomy for the vehicle. The planners also have potential of getting upgraded adding more assumptions about real underwater environment and testing other algorithms to achieve better accuracy and real-time performance.

This study is a part of the "Autonomous Underwater Mission and Exploration Project" conducting at the Centre for Maritime Engineering, Control and Imaging, Flinders University, Australia. As the performance validation of the DEFO model has been completed successfully in this paper, as a future work, the DEFO model will be implemented on-board the under developing Flinders AUVs [11, 21] for experimental evaluation and further investigations.

## Biographies:

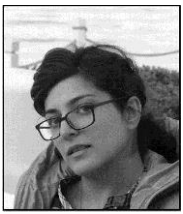

**Somaiyeh MahmoudZadeh** completed her PhD in Computer Science (Robotics and Autonomous Systems) at College of Science and Engineering, Flinders University of South Australia. Her area of research includes computational intelligence, autonomy and decision making, situational awareness, and motion planning of autonomous underwater vehicles.

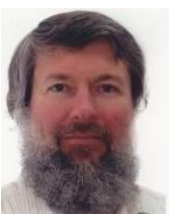

**David M W Powers** is Professor of Computer Science and Director of the Centre for Knowledge and Interaction Technology and has research interests in the area of Artificial Intelligence and Cognitive Science. His specific research framework takes Language, Logic and Learning as the cornerstones for a broad Cognitive Science perspective on Artificial Intelligence and its practical applications. Prof. Powers also serves on several programming committees and editorial boards and being Editor-in-Chief of the Springer journal Computational Cognitive Science and the Springer book series Cognitive Science and Technology.

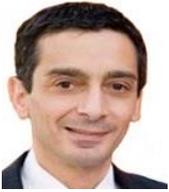

**Karl Sammut** completed his PhD at The University of Nottingham (U.K) in 1992. He is currently an Associate Professor with the Engineering Discipline in the School of Computer Science, Engineering and Mathematics. His areas of expertise are in embedded systems, robotics, smart structures, and maritime electronics. A/Prof Sammut is the Director of the Centre for Maritime Engineering, Control and Imaging at Flinders University and was a Cluster Project leader for the CSIRO Wealth from Oceans Flagship funded project on AUV based pipeline monitoring.

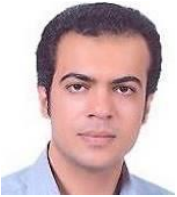

**Amir Mehdi Yazdani** received his PhD in Control Engineering with specialty in Robotics and Autonomous Systems from College of Science and Engineering, Flinders University of South Australia in 2017. His main research interests focus on guidance and control of unmanned vehicles, optimal control and state estimation theory, and intelligent control applications.

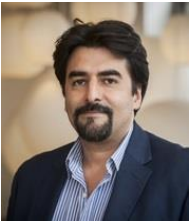

**Adham Atyabi** received his PhD from Flinders University of South Australia in 2013 on "Evolutionary Optimization of Brain Computer Interfaces: Doing More with Less". He was a Postdoctoral Research Associate at Yale Chile Study Center in Yale University. He is currently acting as Technology Lead in Seattle Children's Inovation & Technology Lab and Senior Postdoctoral Fellow at University of Washington. His research interests include Brain Computer Interfacing, EEG analysis, Eye Tracking & Image Processing, Signal Processing, Machine Learning, Swarm and Cognitive Robotics, Knowledge Transfer, and Evolutionary Optimization.